\documentclass[letterpaper, 10 pt, conference]{ieeeconf}  

\IEEEoverridecommandlockouts                              

\usepackage{graphicx}
\usepackage{graphics} 
\usepackage{epsfig} 
\usepackage{mathptmx} 
\usepackage{hhline}
\usepackage{times} 
\usepackage{amsmath} 
\usepackage{cite}

\usepackage{array}
\usepackage{amssymb}  
\usepackage{caption}
\usepackage{float}
\usepackage{tikz}
\usepackage{adjustbox}
\usepackage{bm}
\usepackage{colortbl}
\usepackage{makecell}
\usepackage{nicematrix}
\usepackage{booktabs}
\usetikzlibrary{shapes,arrows}
\usepackage{makeidx}
\makeindex

\usepackage{afterpage}

\title{\LARGE \bf
Exponential Auto-Tuning Fault-Tolerant Control of N Degrees-of-Freedom Manipulators Subject to Torque Constraints*
}

\author{Mehdi Heydari Shahna$^{1}$ and Jouni Mattila$^{2}$
\thanks{*Funding for this research was provided by the Business Finland partnership project "Future All-Electric Rough Terrain Autonomous Mobile Manipulators" (Grant No. 2334/31/2022).}
\thanks{$^{1,}{^{2}}$All authors are with Faculty of Engineering and Natural Sciences, Tampere University, Finland
        {\tt\small \{Mehdi.heydarishahna
Jouni.mattila\}@tuni.fi}}%
}

\begin{document}

\begin{titlepage}
    \centering
    \vspace*{5cm}
    {\Huge \textcolor{red}{This paper has been accepted in: } }\\
    \vspace{1cm}
    {\Huge\textcolor{red}{63rd IEEE Conference on Decision and Control}}\\
    \vspace{1cm}
    {\Huge\textcolor{red}{(CDC)}}
    \vfill
\end{titlepage}

\maketitle
\thispagestyle{empty}
\pagestyle{empty}

\begin{abstract}
This paper presents a novel auto-tuning subsystem-based fault-tolerant control (SBFC) system designed for robotic manipulator systems with n degrees of freedom (DoF). It initially proposes a novel model for joint torques, incorporating an actuator fault correction model to account for potential faults and a mathematical saturation function to mitigate issues related to unforeseen excessive torque. This model is designed to prevent the generation of excessive torques even by faulty actuators. Subsequently, a robust subsystem-based adaptive control strategy is proposed to force system states closely along desired trajectories, while tolerating various actuator faults, excessive torques, and unknown modeling errors. Furthermore, optimal SBFC gains are determined by tailoring the JAYA algorithm (JA), a high-performance swarm intelligence technique, standing out for its capacity to optimize without the need for meticulous tuning of algorithm-specific parameters, relying instead on its intrinsic principles. Notably, this control framework ensures uniform exponential stability (UES).
The enhancement of accuracy and tracking time for reference trajectories, along with the validation of theoretical assertions, is demonstrated through the presentation of simulation outcomes.


\end{abstract}

\section{INTRODUCTION}
Subsystem-based control, when applied to high-degree-of-freedom (DoF) manipulator systems, represents two distinct facets. On the positive side, utilizing various techniques to decompose a complex and high-order system into subsystems can assist in the development of localized control strategies and in the assessment of stability at the subsystem level \cite{zhu2010virtual}--\cite{9891804}.
Conversely, and negatively, a different form of complexity is introduced, arising from modularity, particularly when encountering state- and time-variant uncertainties, as well as failures, which are commonplace in real-world industrial settings \cite{li2022backstepping}--\cite{ren2023fuzzy}. Failures in autonomous and intelligent robotic systems can stem from various events, including internal actuator issues, power supply system failures, or wiring problems \cite{capisani2012manipulator}, impairing their performance, rendering them incapable of carrying out their tasks, and necessitating the design of fault-tolerant control mechanisms to ensure their continued safe operation without causing harm \cite{siqueira2011robust}--\cite{isermann2011fault}. As one potential remedy, many studies have focused on passive fault-tolerant control (PFTC) to maintain operational integrity and safety in applications that lack fault diagnosis and active intervention sections \cite{ke2023uniform}. In their work \cite{8998402}, Van and Ge designed a passive fault-tolerant approach to mitigate the rapid effects of faults for robotic manipulators based on a robust backstepping control integrated with other methods. Likewise, in pursuit of achieving both fast response and high-precision tracking performance, Anjum and Guo in \cite{anjum2021finite}, proposed a PFTC system for robotic manipulators, built upon a fractional-order adaptive backstepping approach.\\
\indent Furthermore, considering the limitations imposed by the magnitude of physical actuators, sensors, and interfacing devices, it becomes imperative to account for control input constraints \cite{he2017adaptive}. Deviating from these constraints can result in the emergence of undesirable vibrations, degradations in system performance, and, in some cases, complete system immobilization \cite{du2008fuzzy}. Nohooji, as outlined in \cite{nohooji2020constrained}, enhanced the robustness of his neural adaptive proportional-integral-derivative (PID) control for manipulators by incorporating considerations of constrained behavior during system operation. Similarly, Yang et al. \cite{yang2019event} developed an online integral reinforcement learning strategy to address the challenges of robust constrained control in nonlinear continuous-time systems.\\
\indent Furthermore, to overcome a formidable challenge for subsystem-based control designers, managing the extensive array of control gains that demand meticulous tuning is imperative, as they exert a distinct impact on the system's transient and steady performance, even when deploying highly effective and top-performing control methodologies. As a promising solution to this challenge, population-based optimization algorithms have gained popularity in recent times due to their efficiency. However, the improper tuning of algorithm-specific parameters can lead to increased computational effort or the attainment of suboptimal local solutions \cite{kashani2022population}. In contrast to most other optimization algorithms that necessitate the fine-tuning of algorithm-specific parameters for updating particle positions, the JAYA algorithm (JA) uniquely relies on its inherent principles to adapt and optimize a wide range of problems \cite{rao2017self}. The JA was developed by Rao \cite{rao2016jaya}, with the primary objective of addressing both constrained and unconstrained optimization problems. It stems from an innovative swarm-based heuristic introduced in the work by Nanda et al. \cite{nanda2009maiden}. Further, in \cite{houssein2021jaya}, Houssein and colleagues conducted an extensive review of renowned optimization algorithms. Their investigation revealed that in the task of function minimization, the JA consistently outperformed these well-established swarm-based algorithms, delivering markedly superior results in terms of both precision and convergence speed.
Interestingly, in \cite{bansal2021single}, Bansal and collaborators explored the capabilities of three distinct optimization algorithms for fundamental backstepping control of a single-link flexible joint manipulator system. Similar to \cite{houssein2021jaya}, their investigative findings indicated that JA optimization consistently outperformed the other methods in terms of fitness value.\\
\indent In consideration of the critical significance of robust control in ensuring both the safety and performance of robotic manipulators, this paper proposes a novel robust adaptive subsystem-based control to maintain the system's uniform exponential stability (UES) while tolerating various actuator faults, excessive torques, and unknown modeling errors. It not only incorporates the management of joint failures but also optimizes the control parameters by customizing the highly promising swarm intelligence technique (JA). Therefore, the present study offers notable contributions to the field of robotics: (1) it introduces an innovative model for joint torques across different types of actuator functions: normal functioning (healthy mode), stuck failure, performance loss (encompassing incipient and abrupt faults), and saturation (excessive torque). Interestingly, this model can be designed to prevent the generation of excessive torques even by faulty actuators; (2) it introduces a novel SBFC approach tailored for robotic manipulators with n DoF, capable of tolerating diverse actuator faults, excessive torques, and unknown modeling errors; (3) to optimize the SBFC gains, a multi-population and single-phase swarm intelligence technique (JA) is amended, standing out for its capacity to optimize without the need for meticulous tuning of algorithm-specific parameters, relying instead on its intrinsic principles;
(4) and the proposed control strategy ensures the achievement of UES.\\
\indent The remaining sections of this paper are organized as follows: In Section II, the conventional model of an n DoF robotic manipulator is augmented with comprehensive actuator fault models and torque signal constraints. Section III outlines the step-by-step design of the SBFC strategy and presents the stability analysis. The effectiveness of the proposed strategy is thoroughly investigated in the final section.
\section{Modeling the System and Defining the Problem}
\subsection{N DoF Manipulator}
Considering the typical robotic manipulator dynamics, as detailed in \cite{nohooji2020constrained}, we have:
\begin{equation}
\small
\begin{aligned}
\label{equation: 1}
\bm{I}(\bm{q}) \bm{\ddot{q}}= &\bm{T}-\bm{C_m}(\bm{q}, \bm{\dot{q}}) \bm{\dot{q}}-\bm{f}(\bm{\dot{q}})-\bm{G}(\bm{q})-\bm{\bm{T}_L} \cdot
\end{aligned}
\end{equation}
In the given context, $\bm{q}\in \mathbb{R}^n$ represents the generalized joint coordinate vector comprising ‘n’ joints. $\bm{I(q)}:\mathbb{R}^n \rightarrow \mathbb{R}^{n\times n}$ characterizes the mass (inertia) properties, while $\bm{C_m(q, \dot{q})} :\mathbb{R}^n \times \mathbb{R}^n \rightarrow \mathbb{R}^{n \times n}$ accounts for the centrifugal and Coriolis forces. $\bm{G(q)}:\mathbb{R}^n \rightarrow \mathbb{R}^n$ represents the gravitational forces/torques, and $\bm{f(\dot{q})}:\mathbb{R}^n \rightarrow \mathbb{R}^n$ accounts for the resistance encountered during movement. The vector $\bm{T}=[T_{1},\ldots,T_{n}]^{\top}$ represents the generalized continuous torque applied at the joints, and $\bm{\bm{T}_L}\in \mathbb{R}^n$ signifies unaccounted-for external disturbances that affect each joint. Notably, the inertia matrix $\bm{I}(\bm{q})$ possesses the properties of being symmetric, positive, and definite; thus, we can also say: 
\begin{equation}
\small
\begin{aligned}
\label{equation: 2}
0<I_{min} (\bm{I}(\bm{q})^{-1}) \leq\|\bm{I}(\bm{q})^{-1}\| \leq I_{max}(\bm{I}(\bm{q})^{-1}) ,
\end{aligned}
\end{equation}
where $\|\cdot\|$ denotes the squared Euclidean norm, $I_{max}(.) \in \mathbb{R}^+$ and $I_{min}(.)\in \mathbb{R}^+$ represent the matrix $\bm{I}(\bm{q})^{-1}$'s maximum and minimum eigenvalues, respectively.\\
\subsection{Passive Faulty Dynamic Model}
Next, we integrate the fault correction functionality into the established control algorithm. To do so, we adopt the following fault model for the actuator \cite{shahna2020anti}:
\begin{equation}
\small
\begin{aligned}
\label{equation: 3}
\bm{T} = \bm{T}_{\bm{c}} + \bm{\epsilon}(\bm{T}_{\bm{s a t}} - \bm{T}_{\bm{c}}),
\end{aligned}
\end{equation}
where $\bm{T}_{\bm{c}}\in \mathbb{R}^n$ represents the normal command control during the system's healthy state. We use $\bm{\epsilon} = \operatorname{diag}(\epsilon_1, \ldots, \epsilon_n)$ and $\bm{T}_{\bm{s a t}} \in \mathbb{R}^n$ to characterize various types of actuator failures, with ${t_f}$ signifying the period of fault occurrence. When $\epsilon_i = 0$, the corresponding actuator is functioning normally. When $T_{s a t(i)} \neq 0$ indicates a stuck failure. Meanwhile, $0 < \epsilon_i < 1$ represents a performance loss. The behavior model of the fault, when $0 < \epsilon_i < 1$, is extended, as follows:
\begin{equation}
\small
\begin{aligned}
\label{equation: 4}
\epsilon_i= 1-e^{-\gamma_i t} \hspace{0.4cm} t \in t_f, \hspace{0.6cm} \gamma_i > 0,
\end{aligned}
\end{equation}
where $\gamma_i$ represents the rate of evolution of an undisclosed fault. A small $\gamma_i$ value indicates slow fault development, termed an incipient fault. Conversely, a high $\gamma_i$ value results in the time course $\gamma_i$ approximating a step form, classified as an abrupt fault \cite{7462227}.\\
\indent \textbf{Remark (1):} In this paper, we assume that $\epsilon_i\neq1$. This assumption is crucial because $\epsilon_i=1$ signifies an uncompensatable fault in the $i$th actuator, resulting in a complete loss of control access. In cases of such severe faults, control strategies become impractical, necessitating the exploration of mechanical alternatives. These alternatives are unrelated to the concept of control strategies. They may involve actions such as replacing or repairing the faulty actuator or introducing an additional actuator to compensate for the control failure, as discussed in \cite{shahna2020anti}.
\subsection{Torque Signal Constraint}
In addition to addressing actuator faults, our objective is to account for the torque constraints to ensure they do not exceed the specified nominal torque values. For $i=1,\ldots,n$ joints to operate in compliance with the constraints imposed on the control torque $T_i(t)$ for each joint, whether in a healthy or faulty state. This is achieved as follows:
\begin{equation}
\small
\begin{aligned}
\label{equation: 6}
 S(T_{i}(t)) = \begin{cases}\bar{T}_{i}, &  T(t) \geq \bar{T}_{i} \\
T(t) &  \underline{T}_{i} \leq T_(t) \leq \bar{T}_{i} \\
\underline{T}_{i} &  T(t) \leq \underline{T}_{i}\end{cases}\cdot
\end{aligned}
\end{equation}
In this context, $\bar{T}_{i}$ and $\underline{T}_{i}$ denote the upper and lower nominal torque bounds, respectively, of the permissible $T_{i}(t)$ values that can be generated. Consequently, we define:
\begin{equation}
\small
\begin{aligned}
\label{equation: 666}
\bm{S(T)}=[S_1(T_1(t)),\ldots,S_n(T_n(t))]^{\top}.
\end{aligned}
\end{equation}
To elaborate further, we define a constraint model as follows:
\begin{equation}
\small
\begin{aligned}
\label{equation: 7}
S_i(T_{i}(t))=s_{1i} \hspace{0.1cm} T_{i}(t) + s_{2i} ,
\end{aligned}
\end{equation}
where
\begin{equation}
\small
\begin{aligned}
\label{equation: 8}
 s_{1i} = \begin{cases} \frac{1}{\mid T_{i}(t) \mid +1}, & \hspace{0.2cm} T_{i}(t) \geq \overline{T}_{i} \hspace{0.2cm}  \text{or} \hspace{0.2cm} T_{i}(t) \leq \underline{T}_{i} \\
1 & \hspace{0.2cm} \underline{T}_{i}\leq T_{i}(t) \leq \bar{T}_{i} \end{cases} 
\end{aligned}
\end{equation}
and
\begin{equation}
\small
\begin{aligned}
\label{equation: 9}
s_{2i}= \begin{cases}\overline{T}_{i} - \frac{T_{i}(t)}{\mid T_{i}(t) \mid + 1}, & T_{i}(t) \geq \overline{T}_{i} \\
0 &  \underline{T}_{i} \leq T_{i}(t) \leq \overline{T}_{i} \\
\underline{T}_{i} - \frac{T_{i}(t)}{\mid T_{i}(t) \mid + 1} & T_{i}(t) \leq \underline{T}_i\end{cases} \cdot
\end{aligned}
\end{equation}
It is evident that Eqs. \eqref{equation: 7}, \eqref{equation: 8}, and \eqref{equation: 9} imply Eq. \eqref{equation: 6}. We have $s_{2i} \leq \max(|\underline{T}_{i}| + 1, |\overline{T}_{i}| + 1)$ and $s_{1i} \leq 1$. In addition, if we generally say $\bm{s}_1=\text{diag}(s_{11},\ldots,s_{1n})$, and $\bm{s}_2=[s_{21},\ldots,s_{2n}]^{\top}$, we can have from \eqref{equation: 3} and \eqref{equation: 666}:
\begin{equation}
\small
\begin{aligned}
\label{equation: 999}
\bm{S}(\bm{T}) = \bm{s_1}\bm{T}+\bm{s_2},
\end{aligned}
\end{equation}
By incorporating both the faulty dynamic model defined in (\ref{equation: 3}) and the torque constraints outlined in (\ref{equation: 999}), we derive a novel model for joint torques that encompasses both actuator faults and torque constraints. This model is designed to prevent the generation of excessive torques by faulty actuators, as detailed below:
\begin{equation}
\small
\begin{aligned}
\label{equation: 99}
\bm{S}(\bm{T}) =& \bm{S}(\bm{T}_{\bm{c}} + \bm{\epsilon}(\bm{T}_{\bm{s a t}} - \bm{T}_{\bm{c}}))=\bm{s_1}\bm{T}_{\bm{c}} + \bm{s_1}\bm{\epsilon}(\bm{T}_{\bm{s a t}} - \bm{T}_{\bm{c}})+\bm{s_2}\\
=&\bm{s}_1(\bm{I}_{n \times n}-\bm{\epsilon})\bm{T}_{\bm{c}} + \bm{s_1} \bm{\epsilon} \bm{T}_{\bm{s a t}}+\bm{s}_2,
\end{aligned}
\end{equation}
where $\bm{I}_{n \times n} \in \mathbb{R}^{n\times n}$ represents the identity matrix. Consequently, the manipulator dynamics described in \eqref{equation: 1}, incorporating the new fault model introduced in \eqref{equation: 99}, can be reformulated as follows:
\begin{equation}
\small
\begin{aligned}
\label{equation: 10}
\bm{\ddot{q}}= \bm{I^{-1}}(\bm{q})[&\bm{s}_1(\bm{I}_{n \times n}-\bm{\epsilon})\bm{T}_{\bm{c}} +\bm{s}_2 -\bm{C_m}(\bm{q}, \bm{\dot{q}}) \bm{\dot{q}}-\bm{f}(\bm{\dot{q}})-\bm{G}(\bm{q})\\
&-\bm{\bm{T}_L}+ \bm{s}_1 \bm{\epsilon} \bm{T}_{\bm{s a t}}] \cdot
\end{aligned}
\end{equation}
For convenience, we can consider: 
\begin{equation}
\small
\begin{aligned}
\label{equation: 11}
&\bm{\bar{\lambda}}=\bm{s}_1(\bm{I}_{n \times n}-\bm{\epsilon})=\text{diag}(\bar{\lambda}_1,\ldots,\bar{{\lambda}}_n),
\hspace{0.2cm} 0 < \bar{\lambda}_i \leq 1\\
&\bm{s_{max}}=\bm{s}_2 + \bm{s}_1 \bm{\epsilon} \bm{T}_{\bm{s a t}} \cdot
\end{aligned}
\end{equation}
where we define a positive constant $\bar{\lambda}_{min}<\inf(\bar{\lambda}_i)$ (see Remark 1). Then, the ultimate expression of the n DoF of a robotic manipulator is, as follows:
\begin{equation}
\small
\begin{aligned}
\label{equation: 12}
\bm{\ddot{q}}= \bm{I^{-1}}[&\bm{\bm{\bar{\lambda}}}\bm{T}_{\bm{c}} +\bm{s}_{max} -\bm{C_m} \bm{\dot{q}} -\bm{f}-\bm{G}-\bm{\bm{T}_L}] \cdot
\end{aligned}
\end{equation}
\section{Design of the SBFC Strategy}
\begin{figure*} [t]
    \centering
    \includegraphics[width=0.9\textwidth, height=6cm]{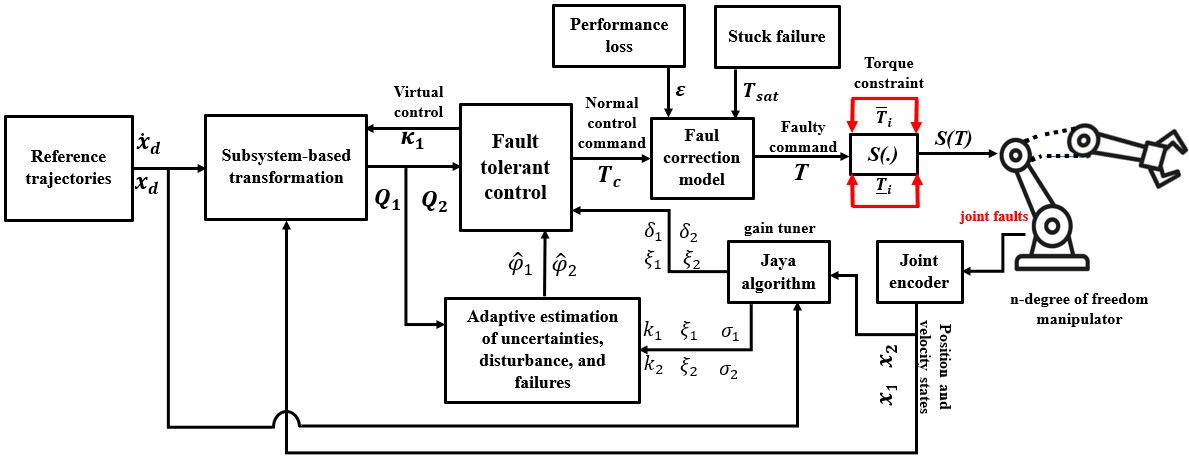}
    \caption{The interconnection among various sections of the proposed control system.}
    \label{fig11}
\end{figure*}
\subsection{Adaptive SBFC Strategy}
To apply the subsystem-based control methodology, the dynamics of a manipulator robot, provided in \eqref{equation: 12}, can be transformed into a triangular feedback form as shown below:
\begin{equation}
\small
\left\{\begin{aligned}
\label{equation: 13}
\dot{\bm{x}}_1(t) & = \bm{x}_2(t) \\
\dot{\bm{x}}_2(t) & =\bm{A}_1 \bm{\bm{\bar{\lambda}}}\bm{T_c}+\bm{g}_1(\bm{x},t) +\bm{\Delta}_1\left(\boldsymbol{x}, t\right)+ \bm{T}_L
\end{aligned}\right. \cdot
\end{equation}
Let us define two state variables $\bm{x}=[\bm{x}_1,\bm{x}_2]^{\top}$, $\bm{\bm{x}}_1=\bm{q} $ as the position vector and $\bm{x}_2=\bm{\dot{q}}$ as the velocity vector. The control torque input incorporates a non-zero coefficient $\bm{A}_1=\bm{I^{-1}}(\bm{q})$. The term $\bm{g}_1(\bm{x},t)$ can be considered established functional elements derived from the system's model, given by $\bm{I^{-1}}(\bm{q})(-\bm{C_m}(\bm{q}, \bm{\dot{q}}) \bm{\dot{q}}-\bm{G}(\bm{q}))$. Meanwhile, $\bm{\Delta}_1(\bm{x},t)$ characterizes uncertain aspects arising from incomplete knowledge of system parameters or modeling inaccuracies, expressed as $\bm{I^{-1}}(\bm{q})(-\bm{f}(\bm{\dot{q}})+\bm{s}_{max})$. In continuation of the preceding form, we can define the tracking error $\bm{e}=[\bm{e_1},\bm{e_2}]^{\top}$, as follows:
\begin{equation}
\small
\begin{aligned}
\label{equation: 14}
 \bm{e}_1=\hspace{+0.1cm}\bm{\bm{x}}_1-\bm{x_{d}}, \hspace{0.2cm}\bm{e}_2=\hspace{+0.1cm}\bm{x}_2-\bm{\dot{x}_{d}},
\end{aligned}
\end{equation}
where $\bm{x_{d}} \in \mathbb{R}^n$ and $\bm{\dot{x}_{d}} \in \mathbb{R}^n$ are the position and velocity reference trajectories, and $\bm{e}_1:\mathbb{R}^n \times \mathbb{R}^n \rightarrow \mathbb{R}^n$ and $\bm{e}_2:\mathbb{R}^n \times \mathbb{R}^n \rightarrow \mathbb{R}^n$ are the position and velocity tracking errors, respectively. Now, we can transform the tracking system into a new form:
\begin{equation}
\small
\begin{aligned}
\label{equation: 15}
 \bm{Q}_1 =\hspace{+0.1cm} \bm{e}_1, \hspace{0.2cm}\bm{Q}_2 =\hspace{+0.1cm} \bm{e}_2 - \bm{\kappa}_{1} \cdot
\end{aligned}
\end{equation}
We introduce the virtual control $\bm{\kappa}_1 \in \mathbb{R}^n$, as follows:
\begin{equation}
\small
\begin{aligned}
\label{equation: 23}
{\bm{\kappa}_{1}} = - \frac{1}{2} (\delta_{1}+\zeta_{1}\hat{\phi}_{1}){\bm{Q}_1},
\end{aligned}
\end{equation}
where $\delta_1$ and $\zeta_{1}$ are positive constants. $\hat{\phi}_1$ is an adaptive function law, which is defined, as follows:
\begin{equation}
\small
\begin{aligned}
\label{equation: 21}
\dot{\hat{\phi}}_{1} &= -{k_1}{\sigma_1}\hat{\phi}_{1}+\frac{1}{2}\zeta_{1}{k_1}\|\bm{Q}_1\|^{2},
\end{aligned}
\end{equation}
where $k_1$, $\zeta_{1}$, and $\sigma_{1}$ are positive constants. By derivative of \eqref{equation: 15} and considering \eqref{equation: 13} and \eqref{equation: 14}, we will have:
\begin{equation}
\small
\begin{aligned}
\label{equation: 17}
&\dot{\bm{Q}}_1= {\bm{Q}_2}+\bm{\kappa}_1 \\
&\dot{\bm{Q}}_2= \bm{A}_1 \bm{\bm{\bar{\lambda}}}\bm{T_c}+\bm{g}_1(\bm{x},t) +\overline{\bm{\Delta}}_1\left(\boldsymbol{x}, t\right)+ \bm{T}_L-\bm{\ddot{x}}_d \cdot\\
\end{aligned}
\end{equation}
where $\bm{\ddot{x}_d} \in \mathbb{R}^n$ can be the desired acceleration of the manipulator robot.
By assuming the function $\bm{\kappa}_{1}$ is smooth, we introduce $\overline{\bm{\Delta}}_{1}$, as follows:
\begin{equation}
\small
\begin{aligned}
\label{equation: 16}
&\overline{\bm{\Delta}}_{1}={\bm{\Delta}_{1}}({\bm{\bm{x}}},t)- \frac{\partial \bm{\kappa}_{1}}{\partial \bm{x_{1}}} \frac{\mathrm{d} \bm{x_{1}}}{\mathrm{~d} t}-\frac{\partial \bm{\kappa}_{1}}{\partial \hat{\phi}_{1}} \frac{\mathrm{d} \hat{\phi}_{1}}{\mathrm{~d} t}\cdot
\end{aligned}
\end{equation}
To prevent the complexity from growing unmanageable, as discussed in Wang et al. \cite{wang2015attitude}, we considered the time derivative of the virtual control to be an element of uncertainty in the system.\\ 
\indent \textbf{Assumption (1)}: There exists a continuously smooth and positive function $r_1: \mathbb{R}^n \rightarrow \mathbb{R}^+$ constrained within the uncertainty bound denoted as $\bm{\bar{\Delta}}_1$. In addition, there are positive parameters $\Omega_1$, $D_{max}$, $\Lambda_1$, and $\bar{g}_{max} \in \mathbb{R}^+$, which may also be unknown such that:
\begin{equation}
\small
\begin{aligned}
\label{equation: 18}
&\|\overline{\bm{\Delta}}_1\| \hspace{0.1cm} \leq \Lambda_1 r_1 \hspace{0.1cm}, \hspace{0.2cm} \|\bm{T}_L\| \hspace{0.1cm} \leq  D_{max}\\
&\|\bm{\ddot{x}}_{d}\| \hspace{0.1cm} \leq \Omega_1,  \hspace{0.2cm} \|\bm{g}_1(\bm{x},t)\| \hspace{0.1cm} \leq  \bar{g}_{max}\cdot
\end{aligned}
\end{equation}
Consequently, the actual control $\bm{T_c}$ is proposed, as follows:
\begin{equation}
\small
\begin{aligned}
\label{equation: 24}
\bm{T_c}=\frac{-1}{2}(\delta_{2}+\zeta_{2}\hat{\phi}_{2}){I_{min}}^{-1}{\bm{Q}_2},
\end{aligned}
\end{equation}
where $\delta_2 $ and $\zeta_{2}$ are positive constants, and we know $I_{min}$ from \eqref{equation: 2}. $\hat{\phi}_2$ is an adaptive function law, which is defined, as follows:
\begin{equation}
\small
\begin{aligned}
\label{equation: 212}
\dot{\hat{\phi}}_{2} &= -{k_2}{\sigma_2}\hat{\phi}_{2}+\frac{1}{2}\zeta_{2}{k_2}\|\bm{Q}_2\|^{2},
\end{aligned}
\end{equation}
where $k_2$, $\zeta_{2}$, and $\sigma_{2}$ are positive constants. Assume the adaptive law errors for \eqref{equation: 21} and \eqref{equation: 212} are $\tilde{\phi}_{j}=\hat{\phi}_{j}-\phi^*_{j}$ for $j=1,2$, considering that there are the positive and unknown constants $\phi^{*}_1$ and $\phi^{*}_2 \in \mathbb{R}^+$ to compensate for the adaptive estimation errors, as follows:
\begin{equation}
\small
\begin{aligned}
\label{equation: 19}
 \hspace{-0.1cm} \phi^*_1=& \zeta_1 ^{-1} \\
 \hspace{-0.1cm}  \phi^*_2=& \zeta_{2}^{-1}[1+2{\bar{\lambda}}_{min}^{-1}(\mu_{1} \Lambda_{1}^2+\nu_{1}{D_{max}^2}+\nu_{{2}}\Omega_1^2+\nu_3 \bar{g}_{max}^2)] \cdot
\end{aligned}
\end{equation}
Apart from $\zeta_1$ and $\zeta_2 \in \mathbb{R}^+$, which are used as control design parameters, all remaining parameters in \eqref{equation: 19} are assumed positive but unknown constants. Now, we can obtain:
\begin{equation}
\small
\begin{aligned}
\label{equation: 22}
&\dot{\tilde{\phi}}_{j}=-{k_j}{\sigma_j}\tilde{\phi}_{j}+\frac{1}{2}\zeta_{j}{k_j}\|{\bm{Q}_j}\|^2-{k_j}{\sigma_j}{\phi}^*_{j} \cdot
\end{aligned}
\end{equation} 
\indent \textbf{Lemma (1)}\cite{heydari2024robust}: According to the general solution of the given linear first-order ordinary differential equations in \eqref{equation: 21} and \eqref{equation: 212}, by choosing an initial condition $\hat{\phi}_j(t_0) > 0$, given that the exponential component of $\hat{\phi}_j$ is monotonically decreasing, and considering the positivity of $k_j$, ${\sigma_j}$, and $\zeta_{j}$, we assert that, for all $t\geq t_0=0$, it is possible to ensure $\hat{\phi}_j(t) > 0$.\\
\indent \textbf{Definition (1)} \cite{corless1993bounded,heydari2024robust}: For any initial condition $\bm{x}(t_0)$, if $\alpha$, $\beta$, and $\tilde{\mu} \in \mathbb{R}^+$ exist, the tracking error $\bm{e}$ between the state $\bm{x}$ and the reference states $\bm{x_r}=[\bm{x_d}, \bm{\dot{x}_d}]^{\top}$ converges uniformly and exponentially to a defined region $g\left(\tau\right)$, such that:
\begin{equation}
\small
\begin{aligned}
\label{equation: 20}
&\|\bm{e}\|=\|\bm{x}(t) - \bm{x_r}(t)\| \leq \beta e^{-\alpha (t-t_0)} \|\bm{x}(t_0)-\bm{x_r}(t_0)\| + \tilde{\mu}\\
&g\left(\tau\right):=\left\{\bm{e} \mid\|\bm{e} \| \leq \tau={\tilde{\mu}}\right\} \cdot
\end{aligned}
\end{equation}
\subsection{JAYA Algorithm-Based Parameter Tuning}
Given the eight gains in the SBFC, denoted as $k_1$, $k_2$, $\delta_1$, $\delta_2$, $\zeta_1$, $\zeta_2$, $\sigma_1$, and $\sigma_2$, it is necessary to tune each within an iterative function based on the multipopulational JA. Let us consider each gain to be associated with $c \in \mathbb{R}^+$. In this paper, the JA commences by initializing two positive collections of gains of control, known as the initial population, through a random process. For each individual within this population, the cost function is calculated, based on the standard deviation of the position and velocity tracking errors $\bar{e}=\sqrt{\|\bm{e}_1\|^2+\|\bm{e}_2\|^2}$ representing the target objective function to be minimized. The top-performing candidate ($c_{best}$) is determined as the one with the most favorable value (referred to as $\bar{e}_{\text{best}}$), while the other (the poorest performer) is identified as the candidate ($c_{worst}$) with the least favorable value (referred to as $\bar{e}_{\text{worst}}$). Next, these values are iteratively adjusted to find the new candidate ($c_{\text{new}}$) in the following iterative function:
\begin{equation}
\small
\begin{aligned}
\label{equation: 25}
c_{new} = c + r_{1}(c_{best} - c) - r_{2}(c_{worst} - c),
\end{aligned}
\end{equation} 
where $c_{\text{new}} \in \mathbb{R}^+$ is the updated random $c$. Further, $r_1$ and $r_2 \in [0,1]$ are the two random numbers for each variable
during the iteration, and $c_{\text{best}}$ and $c_{\text{worst}}$ are replaced with $c_{new}$ if it gives a better $\bar{e}$ than $\bar{e}_{best}$ or worse $\bar{e}$ than $\bar{e}_{worst}$ values, respectively. All accepted function values at the end of the iteration are maintained, and these values become the input to the next iteration.\\
\indent \textbf{Remark (2):} The expression $r_{1}(c_{best}-c)$ represents the inclination of the solution to approach the best solution, while the expression $-r_{2}(c_{worst}-c)$ signifies the propensity of the solution to eschew the worst solution.\\
\indent \textbf{Remark (3):} According to the details provided in this paper, all gain parameters must be both positive and finite. To ensure adherence to this requirement, we must first choose initial populations for these parameters to be positive. Then, by following this approach and referring to \eqref{equation: 25}, while bearing in mind that $c > 0$, we can suggest random $c$ be larger than $\frac{r_2 c_{worst}-r_1 c_{best}}{1-r_1+r_2}$, as well. In this way, we can guarantee that all newly generated values for $c_{new}$ will remain positive. Furthermore, by incorporating the principle constraint outlined in Eq. \eqref{equation: 6}, we can impose constraints on the JA to prevent it from producing gains that exceed a predetermined threshold, as necessary.\\
\indent The block diagram shown in Fig. \ref{fig11} illustrates the interaction among the SBFC system sections. As depicted in the figure, the system computes variables related to subsystem-based transformation upon receiving reference trajectories. In addition, the adaptation mechanisms estimate upper bounds for disturbances, uncertainties, and actuator failures. Then, the calculated values from the subsystem-based transformation component, along with the parameters estimated through online adaptation update laws, are received by the proposed controller. Subsequently, the standard control command, denoted as $\bm{T_c}$, is generated for the novel correction fault model, which incorporates torque constraints. The input constraint section, $\bm{S(T)}$, ensures that both faulty and normal torque values do not surpass the defined constraints. Notably, the eight gains of the adaptation law and controller are automatically adjusted using the JA block in accordance with a cost function that includes the motion sensor data of the robot manipulator and the reference trajectories.
\subsection{Stability Analysis}
\textbf{Theorem (1):} Consider the adaptive algorithm presented in Eq. \eqref{equation: 21}, and \eqref{equation: 212}, the faulty dynamic model for the robotic manipulator in Eqs. \eqref{equation: 12}, and the control input as given in \eqref{equation: 24}. It is assumed that under these conditions, the states $\bm{x_1}$ and $\bm{x_2}$ can attain the reference trajectories $\bm{x_d}$ and $\bm{\dot{x}_d}$ through UES, as defined in Definition (3).\\
\indent \textbf{Proof:} A Lyapunov function is suggested as follows:
\begin{equation}
\small
\begin{aligned}
\label{equation: 26}
&V_{1} =\frac{1}{2} \hspace{0.1cm}\overline{\lambda}_{min} [ \bm{Q}_1^{\top}\bm{Q}_1+{k^{-1}_{1}} \tilde{\phi}_{1}^{2} ] \cdot
\end{aligned}
\end{equation} 
where we define $\bar{\lambda}_{min}$ a positive constant that is less than the infimum of $\bar{\lambda}_i$ provided in \eqref{equation: 11}. After differentiating $V_1$ and inserting \eqref{equation: 17}, we obtain:
\begin{equation}
\small
\begin{aligned}
\label{equation: 27}
\dot{V_{1}}=  \overline{\lambda}_{min} \bm{Q}_1^{\top} [\bm{Q}_2+\bm{\kappa}_{1}]+k_{1}^{-1} \overline{\lambda}_{min}\tilde{\phi}_{1}  \dot{\tilde{\phi}}_{1} \cdot
\end{aligned}
\end{equation} 
By using the Cauchy–Schwarz and the squared Euclidean norm concepts:
\begin{equation}
\small
\begin{aligned}
\label{equation: 28}
\dot{V_{1}}\leq& \frac{1}{2} \overline{\lambda}_{min} (\|\bm{Q}_1\|^{2} + \|\bm{Q}_2\|^{2})+ \overline{\lambda}_{min} (\bm{Q}_1^{\top} \bm{\kappa}_{1}+ k_{1}^{-1} \tilde{\phi}_{1} \dot{\tilde{\phi}}_{1}) \cdot
\end{aligned}
\end{equation} 
Then, by considering the definition of $\phi^*_{1}$ in \eqref{equation: 19}, we achieve:
\begin{equation}
\small
\begin{aligned}
\label{equation: 29}
\dot{V_{1}}\leq& \frac{1}{2} \overline{\lambda}_{min}\|\bm{Q}_2\|^{2} + \frac{1}{2} \overline{\lambda}_{min} \zeta_{1} \phi_{1}^* \|\bm{Q}_1\|^{2} + \overline{\lambda}_{min} \bm{Q}_1^{\top} \bm{\kappa}_{1}\\
&+ k_{1}^{-1} \overline{\lambda}_{min} \tilde{\phi}_{1} \dot{\tilde{\phi}}_{1} \cdot
\end{aligned}
\end{equation} 
Now, by inserting $\dot{\tilde{\phi}}_{1}$ and $\bm{\kappa}_{1}$ from \eqref{equation: 22} and \eqref{equation: 23}, we obtain:
\begin{equation}
\small
\begin{aligned}
\label{equation: 30}
\dot{V_{1}}\leq&   \frac{1}{2} \overline{\lambda}_{min}\|\bm{Q}_2\|^{2} + \frac{1}{2} \overline{\lambda}_{min} \zeta_{1} \phi_{1}^* \|\bm{Q}_1\|^{2}-\frac{1}{2}\overline{\lambda}_{min}{\delta}_{1}{\|\bm{Q}_1\|}^2 \\
&-\frac{1}{2}\overline{\lambda}_{min}\zeta_{1}\hat{\phi}_{1}{\|\bm{Q}_1\|}^2-\overline{\lambda}_{min}{\sigma_{1}}{\tilde{\phi}}_{1}^2+\frac{1}{2}\overline{\lambda}_{min}\zeta_{1}{\|\bm{Q}_1\|}^2\tilde{\phi}_{1}\\
&- \overline{\lambda}_{min}{\sigma_{1}}{\phi}^*_{1}\tilde{\phi}_{1} \cdot
\end{aligned}
\end{equation} 
Because $\tilde{\phi}_{1}=\hat{\phi}_{1}-\phi_{1}^*$:
\begin{equation}
\small
\begin{aligned}
\label{equation: 31}
\dot{V_{1}}\leq& \frac{1}{2} \overline{\lambda}_{min} \|\bm{Q}_2\|^{2} -\frac{1}{2}\overline{\lambda}_{min}{\delta}_{1}{\|\bm{Q}_1\|}^2
-\overline{\lambda}_{min}{\sigma_{1}}{\tilde{\phi}}_{1}^2\\
&-\overline{\lambda}_{min}{\sigma_{1}}{\phi}^*_{1}\tilde{\phi}_{1} \cdot
\end{aligned}
\end{equation} 
After dividing $\overline{\lambda}_{min}{\sigma_{1}}{\tilde{\phi}}_{1}^2$ into $\frac{1}{2}\overline{\lambda}_{min}{\sigma_{1}}{\tilde{\phi}}_{1}^2+\frac{1}{2}\overline{\lambda}_{min}{\sigma_1}{\tilde{\phi}}_{1}^2$, and considering \eqref{equation: 26}, we obtain:
\begin{equation}
\small
\begin{aligned}
\label{equation: 32}
\dot{V_{1}}\leq& -\Psi_{1} V_{1} + \frac{1}{2} \overline{\lambda}_{min} {\|\bm{Q}_2\|}^2 -\frac{1}{2}\overline{\lambda}_{min}{\sigma_{1}}{\tilde{\phi}}_{1}^2\\
&-\overline{\lambda}_{min}{\sigma_{1}}{\phi}^*_{1}\tilde{\phi}_{1},
\end{aligned}
\end{equation} 
where
$\Psi_{1} = \min [{\delta}_{1},\hspace{0.3cm}{k_{1}}\sigma_{1}]$.
As $-\frac{1}{2}{\bar{\lambda}}_{min}{\sigma_{1}}{\hat{\phi}_{1}}^2\leq 0$, we eliminate it and reach:
\begin{equation}
\small
\begin{aligned}
\label{equation: 34}
\dot{V_{1}}\leq& -\Psi_{1} V_{1} + \frac{1}{2} {\bar{\lambda}}_{min} {\|\bm{Q}_2\|}^2 +\frac{1}{2}{\bar{\lambda}}_{min}{\sigma_{1}}{\phi_{1}^*}^2 \cdot
\end{aligned}
\end{equation} 
Likewise, the Lyapunov function $V_2$ is suggested as follows:
\begin{equation}
\small
\begin{aligned}
\label{equation: 35}
&V_{2} = V_{1}+\frac{1}{2}\hspace{0.1cm}[\bm{Q}_2^{\top}\bm{Q}_2+k_{2}^{-1} {\bar{\lambda}}_{min} \tilde{\phi}_{2}^2] \cdot
\end{aligned}
\end{equation} 
By differentiating $V_{2}$ and inserting \eqref{equation: 17}, we obtain:
\begin{equation}
\small
\begin{aligned}
\label{equation: 36}
\dot{V_2}\leq& -\Psi_{1} V_{1} + \frac{1}{2} {\bar{\lambda}}_{min} {\|\bm{Q}_2\|}^2 +\frac{1}{2}{\bar{\lambda}}_{min}{\sigma_{1}}{\phi_{1}^*}^2+k_2^{-1} {\bar{\lambda}}_{min} \tilde{\phi}_2\dot{\tilde{\phi}}_2\\
&+\bm{Q}_2^{\top} [\bm{A}_1\bm{\bar{\lambda}}\bm{T_c}+\bm{g}_1(\bm{x},t)+\overline{\bm{\Delta}}_1\left(\boldsymbol{x}, t\right)+ \bm{T}_L-\bm{\ddot{x}}_d]\cdot
\end{aligned}
\end{equation}
Then, by inserting $\bm{T_c}$ from \eqref{equation: 24}:
\begin{equation}
\small
\begin{aligned}
\label{equation: 37}
\dot{V_2}\leq& -\Psi_{1} V_{1} + \frac{1}{2} {\bar{\lambda}}_{min} {\|\bm{Q}_2\|}^2 +\frac{1}{2}{\bar{\lambda}}_{min}{\sigma_{1}}{\phi_{1}^*}^2\\
&- \frac{1}{2}{\bar{\lambda}}_{min} \delta_{2}\|{\bm{Q}_2}\|^2-\frac{1}{2}{\bar{\lambda}}_{min}\zeta_{2}\hat{\phi}_{2}\|\bm{Q}_2\|^2+\bm{Q}_2^{\top} \bm{g}_1 \\
&+\bm{Q}_2^{\top}  \overline{\bm{\Delta}}_1 -\bm{Q}_2^{\top}  \bm{\ddot{x}}_d+\bm{Q}_2^{\top} \bm{T}_L+k_2^{-1} {\bar{\lambda}}_{min} \tilde{\phi}_2\dot{\tilde{\phi}}_2 \cdot
\end{aligned}
\end{equation}
Now, by assuming that $\mu_1$, $\nu_1$, $\nu_2$, and $\nu_3$ are positive constants, according to Young’s
inequality, we can argue:
\begin{equation}
\small
\begin{aligned}
\label{equation: 38}
&\bm{Q}_2^{\top} \overline{\bm{\Delta}}_1 \hspace{0.1cm} \leq  \hspace{0.1cm} \mu_1 \Lambda_1^2 \|{\bm{Q}_2}\|^2+ \frac{1}{4} \mu_1^{-1} r_1^2 \\
&\bm{Q}_2 ^{\top} \bm{T}_L \hspace{0.1cm} \leq \nu_1 D_{max}^2 \|{\bm{Q}_2}\|^2+ \frac{1}{4} \nu_1^{-1}\\
-&\bm{Q}_2 ^{\top} \bm{\ddot{x}}_d\hspace{0.1cm} \leq \hspace{0.1cm} \nu_2 \Omega_1^2 \|{\bm{Q}_2}\|^2 + \frac{1}{4} \nu_2^{-1} \\
&\bm{Q}_2 ^{\top} \bm{g}_1(\bm{x},t)\hspace{0.1cm} \leq \hspace{0.1cm} \nu_3 \bar{g}_{max}^2 \|{\bm{Q}_2}\|^2 + \frac{1}{4} \nu_3^{-1} \cdot
\end{aligned}
\end{equation}
Because we have $\phi^*_2$ from \eqref{equation: 19}, we can obtain:
\begin{equation}
\small
\begin{aligned}
\label{equation: 39}
\dot{V_2}\leq& -\Psi_{1} V_{1} +\frac{1}{2}{\bar{\lambda}}_{min}{\sigma_{1}}{\phi_{1}^*}^2+ \frac{1}{4} \mu_1^{-1} r_1^2+ \frac{1}{2} \overline{\lambda}_{min} \zeta_{2} \phi_{2}^* \|\bm{Q}_2\|^{2}\\
&- \frac{1}{2}{\bar{\lambda}}_{min}{\delta}_{2}\|{\bm{Q}_2}\|^2 -\frac{1}{2}{\bar{\lambda}}_{min}\zeta_{2}\hat{\phi}_{2}\|\bm{Q}_2\|^2+ \frac{1}{4} \nu_1^{-1} + \frac{1}{4} \nu_2^{-1}\\
&+\frac{1}{4} \nu_3^{-1}+k_2^{-1} {\bar{\lambda}}_{min} \tilde{\phi}_2\dot{\tilde{\phi}}_2 \cdot
\end{aligned}
\end{equation}
In addition, by inserting $\dot{\tilde{\phi}}_2$ from \eqref{equation: 22} into 
\begin{equation}
\small
\begin{aligned}
\label{equation: 40}
\dot{V_2}\leq& -\Psi_{1} V_{1} +\frac{1}{2}\sum_{k=1}^2{\bar{\lambda}}_{min}{\sigma_{k}}{\phi_{k}^*}^2+ \frac{1}{4} \mu_1^{-1} r_1^2- \frac{1}{2}{\bar{\lambda}}_{min}{\delta}_{2}\|{\bm{Q}_2}\|^2\\
&+ \frac{1}{4} \sum_{k=1}^3 \nu_k^{-1}-\overline{\lambda}_{min}{\sigma_{1}}{\tilde{\phi}}_{1}^2-\overline{\lambda}_{min} {\sigma_{1}}{\phi}^*_{1}\tilde{\phi}_{1} \cdot
\end{aligned}
\end{equation}
Like \eqref{equation: 34}, and firm \eqref{equation: 35}, we can obtain:
\begin{equation}
\small
\begin{aligned}
\label{equation: 41}
\dot{V_2}\leq& -\Psi_2 V_2 + \frac{1}{4} \mu_1^{-1} r_1^2+ \frac{1}{4} \sum_{k=1}^3 \nu_k^{-1}+\frac{1}{2} \sum_{k=1}^2 {\bar{\lambda}}_{min}{\sigma_k}{\phi_k^*}^2,
\end{aligned}
\end{equation}
where $\Psi_2 = \min [\Psi_1, \hspace{0.3cm} \bar{\lambda}_{min}{\delta}_2,\hspace{0.3cm}{k_2}\sigma_2]$.
Thus, considering $V=V_2$, we can argue:
\begin{equation}
\small
\begin{aligned}
\label{equation: 43}
V=& \frac{1}{2} {\bar{\lambda}}_{min} \bm{Q}^{\top} \bm{{\Upsilon}} \bm{Q} + \frac{1}{2} {\bar{\lambda}}_{min} \bm{\tilde{\phi}}^{\top} \bm{\bm{K}}^{-1} \bm{\tilde{\phi}},
\end{aligned}
\end{equation}
where:
\begin{equation}
\small
\begin{aligned}
\label{equation: 44}
\bm{Q} &= \begin{bmatrix}
    \bm{Q}_1 \\
    \bm{Q}_2 \\
\end{bmatrix}, \hspace{0.2cm}
\bm{{\Upsilon}}=\begin{bmatrix}
    1 & 0 & \\
    0 & \bar{\lambda}^{-1}_{\text{min}} &  \\
    \end{bmatrix},\\
    \bm{\tilde{\phi}} &= \begin{bmatrix}
    \tilde{\phi}_1 \\
    \tilde{\phi}_2 \\
\end{bmatrix}, \hspace{0.2cm}
\bm{K}^{-1} = \begin{bmatrix}
    k_1^{-1} & 0 \\
    0 & k_2^{-1} & \\
\end{bmatrix} \cdot
\end{aligned}
\end{equation}
Thus, according to \eqref{equation: 41}, we obtain:
\begin{equation}
\small
\begin{aligned}
\label{equation: 45}
\dot{V}\leq& -\Psi_2 V+ \frac{1}{4} \mu_1^{-1} r_1^2 + \tilde{\mu},
\end{aligned}
\end{equation}
where
$\tilde{\mu}=\frac{1}{4} \sum_{k=1}^3 \nu_k^{-1}+\frac{1}{2} \sum_{k=1}^2 {\bar{\lambda}}_{min}{\sigma_k}{\phi_k^*}^2$.
In this section, we should recall the following solution (see Lemma 1):
\begin{equation}
\small
\begin{aligned}
\label{equation: 47}
&\dot{V} =\Psi V+\mu r\\
&V=e^{\Psi t} V(0)+\int_0^t e^{\Psi(t-\tau)} \mu r(\tau) d \tau  \cdot
\end{aligned}
\end{equation}
In the same way, we can solve \eqref{equation: 45}, as follows:
\begin{equation}
\small
\begin{aligned}
\label{equation: 48}
V \leq&  V\left(t_0\right) e^{-\left\{\Psi_2\left({t-t_0}\right)\right\}} +\frac{1}{4}  \mu_1^{-1} \int_{t_0}^t e^{\left\{-\Psi_2(t-T)\right\}} r_1^2 \hspace{0.2cm} dT \\
&+{\tilde\mu}\int_{t_0}^t e^{\left\{-\Psi_2(t-T)\right\}}\hspace{0.2cm} dT   \cdot
\end{aligned}
\end{equation}
Considering \eqref{equation: 35}, we know:
$\frac{1}{2}\hspace{0.1cm}\bm{Q}_2^{\top}\bm{Q}_2 \leq V_{2}=V$.
Thus, we can interpret \eqref{equation: 48} as follows:
\begin{equation}
\small
\begin{aligned}
\label{equation: 49}
\|\bm{Q}\|^2 \leq & 2  V\left(t_0\right) e^{-\left\{\Psi_2\left({t-t_0}\right)\right\}} +\frac{1}{2} \mu_1^{-1} \int_{t_0}^t e^{\left\{-\Psi_2(t-T)\right\}} r_1^2\hspace{0.2cm}dT \\
 & + 2\hspace{0.1cm}{\tilde\mu} \hspace{0.1cm} {\Psi_2}^{-1} \cdot
\end{aligned}
\end{equation}
Both $\mu_1$, and $\Psi_2$ are positive constants dependent on designable control gains that are freely chosen to satisfy the following condition:
$\frac{1}{2} < {\mu_1 \Psi_2}$.
To continue the stability proof, we define a continuous operator $Z(\cdot)$, as follows:
\begin{equation}
\small
\begin{aligned}
\label{equation: 51}
Z(\iota)=\frac{1}{2}  \frac{1}{\mu_1(\Psi_2-\iota)}>0, \hspace{0.2cm} \iota \in [0,\Psi_2)  \cdot
\end{aligned}
\end{equation}
It is evident that by increasing $\iota$, $Z(\iota)$ increases, meaning that:
$Z(\iota)\geq Z(0)=\frac{1}{2}  \frac{1}{{\mu_1 \Psi_2}}$.
Because $\mu_1$ is a positive constant, it becomes evident that we can find a small positive value $\bar{\iota} \in \iota$ which $\bar{\iota}\mu_1 < \frac{1}{2}$. By knowing that $\frac{1}{2} < {\mu_1 \Psi_2}$, we can obtain:
$0 < \bar{\iota} < \frac{1}{2\mu_1}$.
Then, we can obtain:
$0 < Z(\bar{\iota})=\frac{1}{2}\frac{\mu_1^{-1}}{\Psi_2-\bar{\iota}} < 1$.
If we say $\bar{Z}=Z(\bar{\iota})$, by multiplying $e^{\bar{\iota}(t-t_0)}$ by \eqref{equation: 49}, we reach:
\begin{equation}
\small
\begin{aligned}
\label{equation: 53}
 {\|\bm{Q}\|^2}e^{\bar{\iota}(t-t_0)} \leq& 2 V(t_0) e^{{-(\Psi_2-\bar{\iota})(t-t_0)}}+2 \tilde{\mu} \Psi_2^{-1} e^{{\bar{\iota}(t-t_0)}}\\
 &+\frac{1}{2} \mu_1^{-1} \int_{t_0}^t e^{{-\Psi_2(t-T)+\bar{\iota}(t-t_0)}} r_1^2 \hspace{0.2cm} dT  \cdot
\end{aligned}
\end{equation}
Because $0 \leq \bar{\iota} <\Psi_2$, we can eliminate the decreasing element $e^{{-(\Psi_2-\bar{\iota})(t-t_0)}}$ from \eqref{equation: 53}:
\begin{equation}
\small
\begin{aligned}
\label{equation: 54}
 {\|\bm{Q}\|^2}e^{\bar{\iota}(t-t_0)} \leq& 2  V(t_0)+2  \tilde{\mu}\Psi_2^{-1} e^{{\bar{\iota}(t-t_0)}} \\
 +&\frac{1}{2} \mu_1^{-1} \int_{t_0}^t e^{-(\Psi_2-\bar{\iota})(t-T)}r_1^2 e^{{\bar{\iota}(t-t_0)}}\hspace{0.1cm} dT  \cdot
\end{aligned}
\end{equation}
We represent the non-decreasing and continuous functions $E_0$ and $E_1$, as follows:
\begin{equation}
\small
\begin{aligned}
\label{equation: 55}
E_0 (t) = & \sup _{\omega \in[t_0,t]} \hspace{0.2cm} [\|\bm{Q}(\omega)\|^2 e^{{\bar{\iota}(\omega-t_0)})}] \\
E_1 (t)= & \sup _{\omega \in[t_0,t]} \hspace{0.2cm} [(r_1^2)  e^{\bar{\iota}(\omega-t_0)}]  \cdot\\
\end{aligned}
\end{equation}
Then, by considering Eqs. \eqref{equation: 54} and \eqref{equation: 55}, we achieve:
\begin{equation}
\small
\begin{aligned}
\label{equation: 56}
 {\|\bm{Q}\|^2}e^{\bar{\iota}(t-t_0)} \leq& 2 V(t_0)+\frac{1}{2}  \frac{\mu_1^{-1}}{\Psi_2-\bar{\iota}} E_1+2 \tilde{\mu} \Psi_2^{-1} e^{{\bar{\iota}(t-t_0)}} \cdot
\end{aligned}
\end{equation}
Because $E_1$ is non-decreasing, the left-hand side of Eq. \eqref{equation: 56} will also not decrease. Hence, with respect to the definition of $E_0$ in Eq. \eqref{equation: 55}, we can conclude:
\begin{equation}
\small
\begin{aligned}
\label{equation: 57}
E_0 \leq & 2  V(t_0)+\frac{1}{2} \frac{\mu_1^{-1}}{\Psi_2-\bar{\iota}} E_1+2 \tilde{\mu} \Psi_2^{-1} e^{\bar{\iota}(t-t_0)} \cdot
\end{aligned}
\end{equation}
By defining
$E=\max {(E_0,E_1)}$, we can obtain:
\begin{equation}
\begin{aligned}
\label{equation: 59}
E_0 \leq 2  V\left(t_0\right)+\bar{Z} E+2 \tilde{\mu} \Psi_2^{-1} e^{\bar{\iota}(t-t_0)},
\end{aligned}
\end{equation}
such that both $E_0$ and $E$ are not decreasing. By choosing $\Psi_2$ large enough, which relies on control gains, and selecting $\bar{\iota}$ small enough, it becomes possible to ensure the existence of a sufficiently large $\Psi_2>\iota^*>\bar{\iota}$ such that $\overset{*}{Z}=Z(\iota^*)$ and satisfy the following condition \cite{heydari2024robust}:
\begin{equation}
\small
\begin{aligned}
\label{equation: 60}
\overset{*}{Z}>\bar{Z} \hspace{0.2cm},\hspace{0.2cm} 0<\overset{*}{Z}<1 \implies \bar{Z} E \leq \overset{*}{Z} E_0  \cdot
\end{aligned}
\end{equation}
\eqref{equation: 60} is justified when we have ${\iota}^* \geq \frac{E_0}{E} \psi_{2} -(\bar{\iota}+\psi_{2})$ which is straightforward.
Incorporating Eq. \eqref{equation: 60} into Eq. \eqref{equation: 59}, we arrive at:
\begin{equation}
\small
\begin{aligned}
\label{equation: 61}
E_0 \leq 2 V(t_0)+\overset{*}{Z} E_0(t)+2 \bar{\mu} \Psi_2^{-1} e^{\bar{\iota}(t-t_0)}  \cdot
\end{aligned}
\end{equation}
Afterward, we obtain:
\begin{equation}
\small
\begin{aligned}
\label{equation: 62}
E_0 \leq \frac{2 V\left(t_0\right)+2 \tilde{\mu} \Psi_2^{-1} e^{\bar{\iota}(t-t_0)}} {1-\overset{*}{Z}}  \cdot
\end{aligned}
\end{equation}
Concerning the definition \eqref{equation: 55}, we obtain:
\begin{equation}
\small
\begin{aligned}
\label{equation: 630}
\|\bm{Q}\|^2 \leq \frac{2 V\left(t_0\right) e^{-\bar{\iota}(t-t_0)}+2 \tilde{\mu} \Psi_2^{-1}} {1-\overset{*}{Z}}  \cdot
\end{aligned}
\end{equation}
It is significant that:
\begin{equation}
\small
\begin{aligned}
\label{equation: 64}
\sup _{t \in\left[t_0, \infty\right]}(\frac{2 V(t_0) e^{-\bar{\iota}(t-t_0)}} {1-\overset{*}{Z}})\leq \frac{2 V(t_0)} {1-\overset{*}{Z}}  \cdot
\end{aligned}
\end{equation}
Consequently, by Definition (1), it is evident from Eq. \eqref{equation: 630} that $||\bm{Q}||$ is uniformly and exponentially stable towards a specific ball $\mathcal{G}\left(\bar{\tau}_0\right)$ when employing the SBFC approach, such that:
\begin{equation}
\small
\begin{aligned}
\label{equation: 65}
\mathcal{G}\left(\bar{\tau}_0\right):=\left\{\bm{Q} \mid\|\bm{Q}\| \leq \bar{\tau}_0=\sqrt{\frac{2 \tilde{\mu} \Psi_2^{-1}}{1-\overset{*}{Z}}}\right\}  \cdot
\end{aligned}
\end{equation}
\begin{figure*} [h]
    \centering
    \includegraphics[width=0.8\textwidth, height=3.8cm]{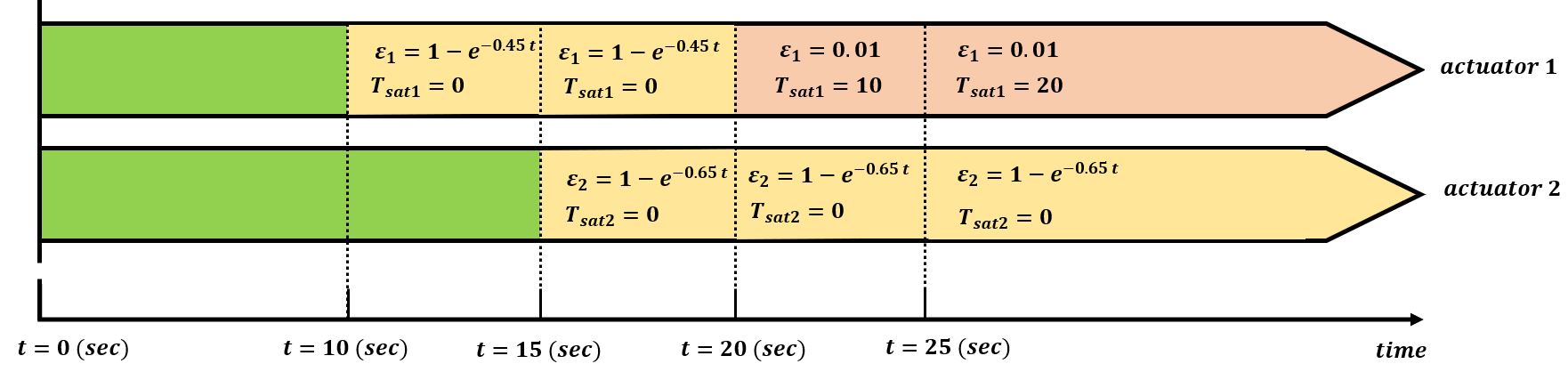}
    \caption{Actuator fault occurrence (seconds): normal: $[0,10]$; one faulty actuator: $[10,15]$, and two faulty actuators: $[15,\infty]$. }
     \label{fig2f}
\end{figure*}
\begin{table}[h!]
  \centering
  \renewcommand{\arraystretch}{0.1} 
  \colorbox{white!10}{\fbox{%
    \begin{tabular}{p{0.88\linewidth}} 
      \multicolumn{1}{c}{\textbf{Algorithm 1}: SBFC approach} \\
      \\
      \hspace{0.4cm}\textbf{Input}: states $\bm{x}$, and references $\bm{x_r}$.  \\
      \hspace{0.4cm}\textbf{Output}: control input $\bm{S(T)}$.\\
      \\
      {\small 1}\hspace{0.7cm}Initialize some $c \in \mathbf{R}^+$;\\
      {\small 2}\hspace{0.7cm}$\bm{e_1}=\bm{x_1}-\bm{x_d}$;\\
      {\small 3}\hspace{0.7cm}$\bm{Q_1}=\bm{e_1}$;\\
      {\small 4}\hspace{0.7cm}$\dot{\hat{\phi}}_{1} = -k_1 \sigma_1\hat{\phi}_{1}+\frac{1}{2}\zeta_{1} k_1\|\bm{Q_1}\|^2$;\\
      {\small 5}\hspace{0.7cm}$\bm{\kappa_1}=-\frac{1}{2}(\delta_{1}+\zeta_{1}\hat{\phi}_{1})\bm{Q_1}$;\\
      {\small 6}\hspace{0.7cm}$\bm{e_2}=\bm{x_2}-\bm{\dot{x}_d}$;\\
      {\small 7}\hspace{0.7cm}$\bm{Q_2}=\bm{e_2}-\bm{\kappa_1}$;\\
      {\small 8}\hspace{0.7cm}$\dot{\hat{\phi}}_{2} = -k_2 \sigma_2\hat{\phi}_{2}+\frac{1}{2}\zeta_{2} k_2\|\bm{Q_2}\|^2$;\\
      {\small 9}\hspace{0.7cm}$\bm{T_c}=-\frac{1}{2}(\delta_{1}+\zeta_{1}\hat{\phi}_{1})\lambda_{min}^{-1}\bm{Q_1}$;\\
      {\small 10}\hspace{0.55cm}$\bm{S(T)}=\bm{s_1} \bm{T} + \bm{s_2}$\\
      {\small 11}\hspace{0.55cm}Find $c_{worst}$, $c_{best}$;\\
      {\small 12}\hspace{0.50cm}Value randomly $c$ following Remark (3);\\
      {\small 13}\hspace{0.55cm}$\bm{c_{new}}=\bm{c}+r_{1}(\bm{c_{best}}-\|\bm{c}\|)-r_{2}(\bm{c_{worst}}-\|\bm{c}\|)$;\\
      {\small 14}\hspace{0.55cm}Repeat the steps from step 2 onward.\\
    \end{tabular}%
  }}
\end{table}
\section{Numerical validity}
\begin{figure}[h!]
    \centering
    \scalebox{0.7}{\includegraphics[trim={0cm 0.0cm 0.0cm 0cm},clip,width=\columnwidth]{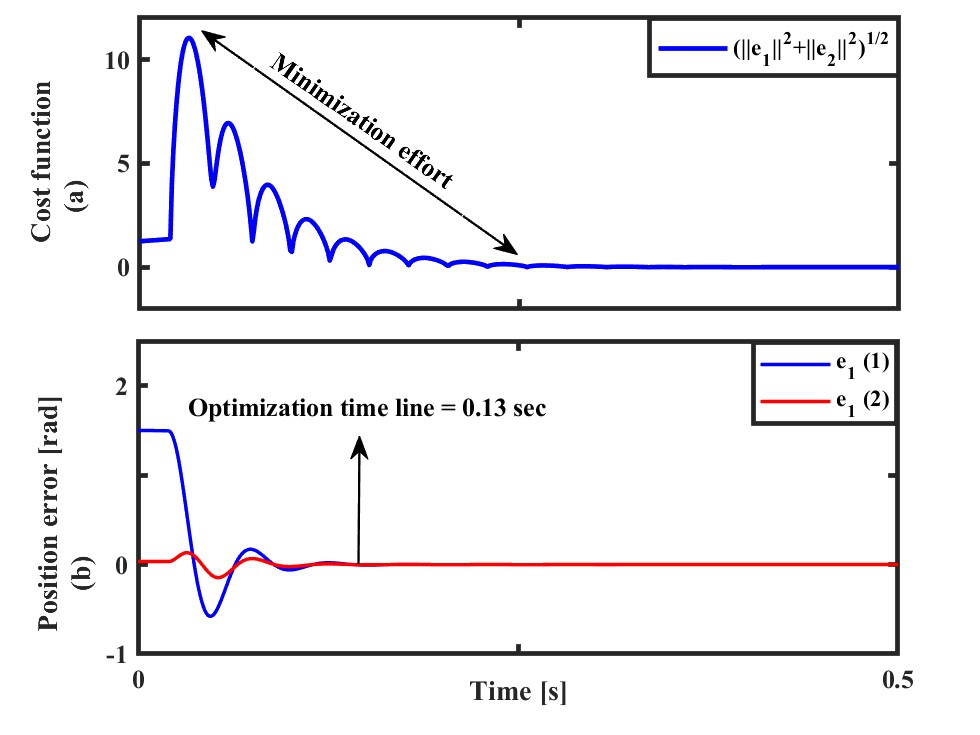}}
    \caption{Cost function (a) and tracking position error (b)}
    \label{fig22}
\end{figure}

\begin{figure}[h!]
    \centering
    \scalebox{0.7}{\includegraphics[trim={0cm 0.0cm 0.0cm 0cm},clip,width=\columnwidth]{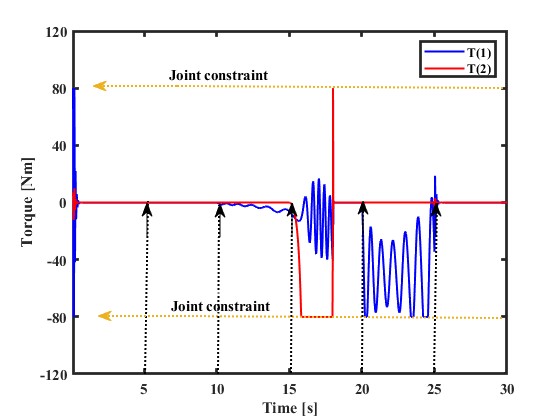}}
    \caption{Torque effort generated by healthy and faulty joints. }
    \label{fig352}
\end{figure}

\begin{figure}[h!]
    \centering
    \scalebox{0.8}{\includegraphics[trim={0cm 0.0cm 0.0cm 0cm},clip,width=\columnwidth]{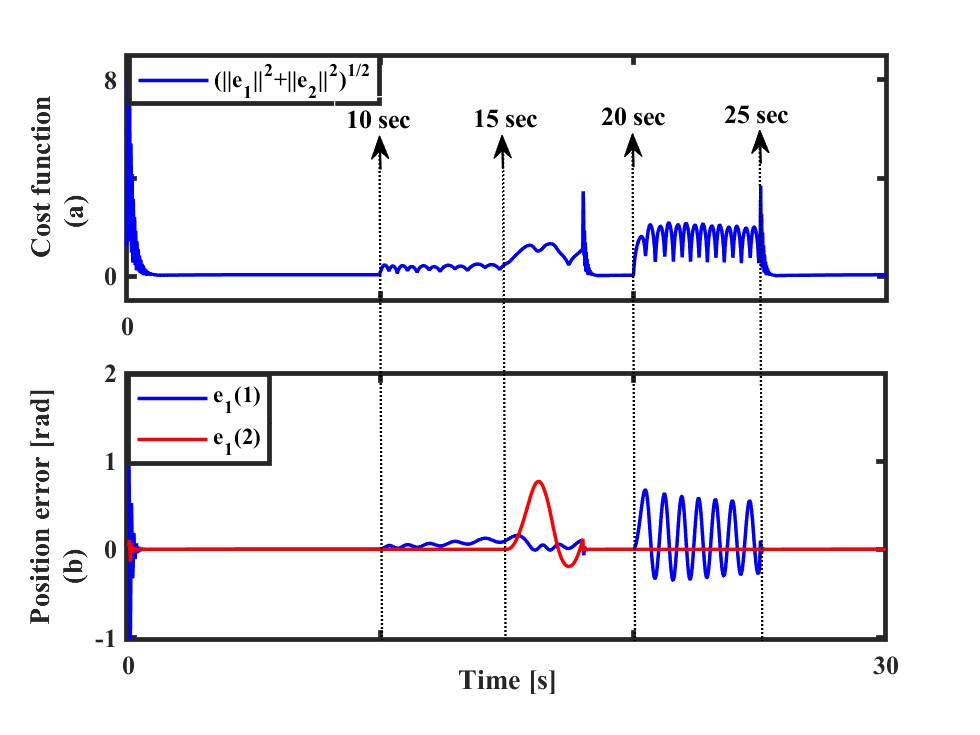}}
    \caption{Cost function (a) and tracking position error (b) }
    \label{fig35}
\end{figure}

The deployment approaches for the SBFC are delineated within Algorithm 1, offering a detailed overview. To evaluate the efficacy of the proposed methodology, we applied it to the 2-DoF vertical
plane robot featured in the work by Humaloja et al. \cite{humaloja2021decentralized} with link lengths 1m and 0.8m, which was based on \cite{10.5555/560653}. The robotic manipulator dynamic and the SBFC strategy were implemented in the simulation, running at a 10-KHz frequency. 
The modeling of the unknown friction and external disturbance term is represented as follows:
\begin{equation}
\small
\begin{aligned}
\label{equation: 3007}
\bm{\Delta_1}+\bm{T_L}=\left[\begin{array}{l}
0.6 \sin \left(0.8\dot{q}_1 q_2\right) +3 \sin(2t) \\
-1.6 \sin \left(1.8 q_2\right)+1.3 \sin \left(0.7\dot{q}_2\right) -0.2 q_2
\end{array}\right]  \cdot
\end{aligned}
\end{equation}
The system's desired trajectory, based on radians, is chosen as follows:
\begin{equation}
\small
\begin{aligned}
\label{equation: 30007}
\bm{x_d}=[\sin (t / 4 \pi)-1, \sin (t / 4 \pi+\frac{\pi}{3})]^T  \cdot
\end{aligned}
\end{equation}
In this case study, we examined a fault model that occurred in both actuators, as represented in Fig. \ref{fig2f}, where both actuators were initially in healthy and normal condition for up to 10 seconds. The effectiveness of JA and tracking control during this healthy task is illustrated in Fig. \ref{fig22}. The JA commenced by initializing two positive collections of gains of control, through a random process. The step time for updating the candidates of the JA was set at 0.0001 seconds. This means that during each step, the control gains are updated to converge to the optimal values. The best SBFC gains obtained at $0.13$ seconds for the mentioned manipulator and the specified task are as follows: $\delta_1=62$, $\delta_2=75$, $\zeta_1=0.2$, $\zeta_2=3.5$, $\sigma_1=5.6$, $\sigma_2=1.9$, $k_1=1.4$, and $k_2=0.96$. This depiction indicates that control parameters were suitably optimized, leading to the cost function in Fig. \ref{fig22}(a) reaching a minimum value at $0.25$ sec. Fig. \ref{fig22}(b) also illustrates the potential for position tracking using SBFC, in the presence of uncertainties, in the healthy actuator mode, in which the position error reached $0.0009$ radian, following parameter tuning.
\\
Fig. \ref{fig352} illustrates the torque efforts generated by healthy and faulty joints to maintain control performance. The torque effort significantly decreased until 10 seconds before faults occurred (see Fig. \ref{fig2f}). The first actuator experiences a variety of faults after 10 seconds, while faults in the second actuator begin at 15 seconds, although each actuator fault affects the other actuator as well. This results in the generation of more frequent torque to compensate for faults. The severe faults for the first and second actuators occur at 20 and 15 seconds, respectively. It demonstrates the effectiveness of the designed constraints, even for faulty actuators, in preventing torques larger than the defined nominal values (80Nm). Despite these challenges, control performance is successfully maintained for around 26 seconds. This is evidenced by the fact that, although the last faults persist without further change, the adaptation control efforts generated by the actuators decrease significantly, ultimately reaching the control goal.
Fig. \ref{fig35} illustrates the system's response to the fault model mentioned in Fig. \ref{fig2f} in terms of the objective function and position tracking error. This demonstrates its capability to effectively reduce tracking errors to zero, even in the presence of faults in both actuators.
\\
\indent Table I compares the performance of SBFC with two similar works \cite{humaloja2021decentralized} with semi-global exponential stability (SGES) and \cite{su2020new} with Finite-time stability (FTS) under identical tasks: healthy actuators, and scenarios with one and two faulty actuators, while all tasks subject to uncertainties as defined in \eqref{equation: 3007} and reference trajectory \eqref{equation: 30007}. We attempted to obtain optimal values for \cite{humaloja2021decentralized} and \cite{su2020new} using Teaching Learning Based Optimization (TLBO). Both TLBO and JA don’t require specific algorithmic parameters. However, in contrast to JA, which has one phase, TLBO consists of two phases, making JA straightforward to implement.
The table demonstrates SBFC's superior performance in terms of tracking control accuracy and speed, especially in faulty scenarios.
\begin{table}[h]
  \captionsetup{position=top}
  \caption{Control performance of the SBFC, \cite{humaloja2021decentralized} and \cite{su2020new} in tracking the joint position reference under various actuators' statuses.}
  \centering
  \scriptsize
  \begin{tabular}{cccccc}
    \toprule
    \textcolor{black}{\textbf{Actuators'}} &\textcolor{black}{\textbf{Convergence}} & \textcolor{black}{\textbf{SBFC}} &
    \textcolor{black}{\textbf{\cite{humaloja2021decentralized}}} &
    \textcolor{black}{\textbf{\cite{su2020new}}} & \\
        \textcolor{black}{\textbf{status}} &\textcolor{black}{\textbf{criteria}} & \textcolor{black}{\textbf{approach}} &
    \textcolor{black}{\textbf{approach}} &
    \textcolor{black}{\textbf{approach}} & \\
    \midrule
    \textcolor{black}{\textbf{Generic}} &\textcolor{black}{\textbf{Stability}} & \textbf{UES} &  \textbf{SGES} & \textbf{FTS} & \\
    \textcolor{black}{\textbf{Normal}} &\textcolor{black}{\textbf{Tracking error}} \textbf{(rad)} & \bm{$0.0009$} & \bm{$0.0025$} & \bm{$0.006$} & \\
   \textcolor{black}{\textbf{Normal}} & \textcolor{black}{\textbf{Tracking speed}} \textbf{(sec)} & \bm{$0.13$} & \bm{$0.85$} & \bm{$0.7$} & \\
      \textcolor{black}{\textbf{one-faulty}} & \textcolor{black}{\textbf{Tracking error}} \textbf{(rad)} & \bm{$0.0012$} & \bm{$0.0031$} & \bm{$0.0071$} & \\
   \textcolor{black}{\textbf{one-faulty}} & \textcolor{black}{\textbf{Tracking  speed}} \textbf{(sec)}  & \bm{$1.1$} & \bm{$2.81$} & \bm{$2.2$} & \\
       \textcolor{black}{\textbf{two-faulty}} & \textcolor{black}{\textbf{Tracking error}} \textbf{(rad)} & \bm{$0.0018$} & \bm{$0.0033$} & \bm{$0.0074$} & \\
    \textcolor{black}{\textbf{two-faulty}} &\textcolor{black}{\textbf{Tracking speed}} \textbf{(sec)}  & \bm{$1.22$} & \bm{$2.93$} & \bm{$2.35$} & \\
    \bottomrule
  \end{tabular}
\end{table}
\section{CONCLUSIONS}
This study introduced a novel robotic manipulator joint model, including different types of actuator functions: normal functioning (healthy mode), stuck failure, performance loss (encompassing incipient and abrupt faults), and saturation (excessive torque). In addition, a subsystem-based fault-tolerant control system tailored to address uniformly exponential stability among robotic manipulator systems, effectively managing unknown modeling errors, external disturbances, and failures, while optimizing controller gains through amending the JA. It stood out for its capacity to optimize without the need for meticulous tuning of algorithm-specific parameters. Looking forward, this generic control methodology opens new avenues for its application across a spectrum of robotic dynamics, suggesting broader implications for future research.
\bibliographystyle{IEEEtran}
\bibliography{ref}

\end{document}